\documentclass[letterpaper, 10pt, conference]{ieeeconf}

\IEEEoverridecommandlockouts
\usepackage{graphics}

\usepackage{amsmath}
\usepackage{amssymb}
\usepackage[colorlinks, linkcolor=blue, citecolor=blue]{hyperref}
\usepackage{cite}

\usepackage{algorithm}
\usepackage{url}
\usepackage{algpseudocode}
\usepackage{epsfig}
\usepackage{graphicx}
\usepackage{caption}
\usepackage{subfigure}
\usepackage{mathrsfs}
\usepackage{array}
\usepackage{multirow}
\usepackage{multicol}
\usepackage{booktabs}
\usepackage{makecell}
\usepackage{xcolor}
\usepackage{amsfonts}
\usepackage{gensymb}
\usepackage{threeparttable}

\title{\LARGE \bf
Socially-Aware Multi-Agent Following with 2D Laser Scans \\ via Deep Reinforcement Learning and Potential Field
}

\author{Yuxiang Cui, Xiaolong Huang, Yue Wang, Rong Xiong
\thanks{Yuxiang Cui,  Xiaolong Huang, Yue Wang and Rong Xiong are with the State Key Laboratory of Industrial Control and Technology, Zhejiang University, Hangzhou, P.R. China. Rong Xiong is the corresponding author.{\tt\small rxiong@zju.edu.cn}.}%
}

\begin{document}

\maketitle
\thispagestyle{empty}
\pagestyle{empty}

\begin{abstract}
Target following in dynamic pedestrian environments is an important task for mobile robots. 
However, it is challenging to keep tracking the target while avoiding collisions in crowded environments, especially with only one robot.
In this paper, we propose a multi-agent method for an arbitrary number of robots to follow the target in a socially-aware manner using only 2D laser scans.
The multi-agent following problem is tackled by utilizing the complementary strengths of both reinforcement learning and potential field, in which the reinforcement learning part handles local interactions while navigating to the goals assigned by the potential field.
Specifically, with the help of laser scans in obstacle map representation, the learning-based policy can help the robots avoid collisions with both static obstacles and dynamic obstacles like pedestrians in advance, namely socially aware.
While the formation control and goal assignment for each robot is obtained from a target-centered potential field constructed using aggregated state information from all the following robots.
Experiments are conducted in multiple settings, including random obstacle distributions and different numbers of robots.
Results show that our method works successfully in unseen dynamic environments. The robots can follow the target in a socially compliant manner with only 2D laser scans. 

\end{abstract}

\section{Introduction}

Robot following in a robust and safe manner is in demand for robots working in crowded environments.
The robots should be able to keep following the target person to perform tasks like security surveillance and physical assistance.
In the process, the robots should make decisions considering the evolution of surrounding dynamics in case of losing the target person or endangering self-safety or human-safety. 
Thus, a socially-aware following method of better stability is needed.

Traditional solutions turn the task of following into a problem of path planning.
They use hierarchical pipelines that assign a goal position around the target first and then plan a collision-free path with simple path planners like $A^*$ \cite{hart1968formal}\cite{honig2018toward} or optimization-based planners using specially designed field. \cite{shin2020optimization}.
These methods usually need a pre-built global map for path planning, and therefore can not be directly transferred into environments with dynamic obstacles. Updating the map in real-time with detection modules can address this problem, but those modules also introduce additional constraints like detection accuracy and robustness.

Deep learning based methods try to tackle the person following problem in an end-to-end style\cite{pang2020efficient}. 
They use deep neural networks to process robots' self-states and raw sensor data like RGB-D images and then directly output the steering commands. 
These methods can be roughly divided into two categories, imitation learning methods and reinforcement learning methods. 
Imitation learning methods optimize the policy over expert demonstrations and obtain the final policy that makes decisions like the experts.
While reinforcement learning methods acquire policy by interacting with the environment, which can be used to learn arbitrary policies given designed reward settings. 
These methods can achieve much more sophisticated behaviors than the traditional methods, but are mainly used in single robot scenarios as the difficulty of policy training increases rapidly with the expansion of robots.

\begin{figure}[t]
\centering
\includegraphics[scale=0.30]{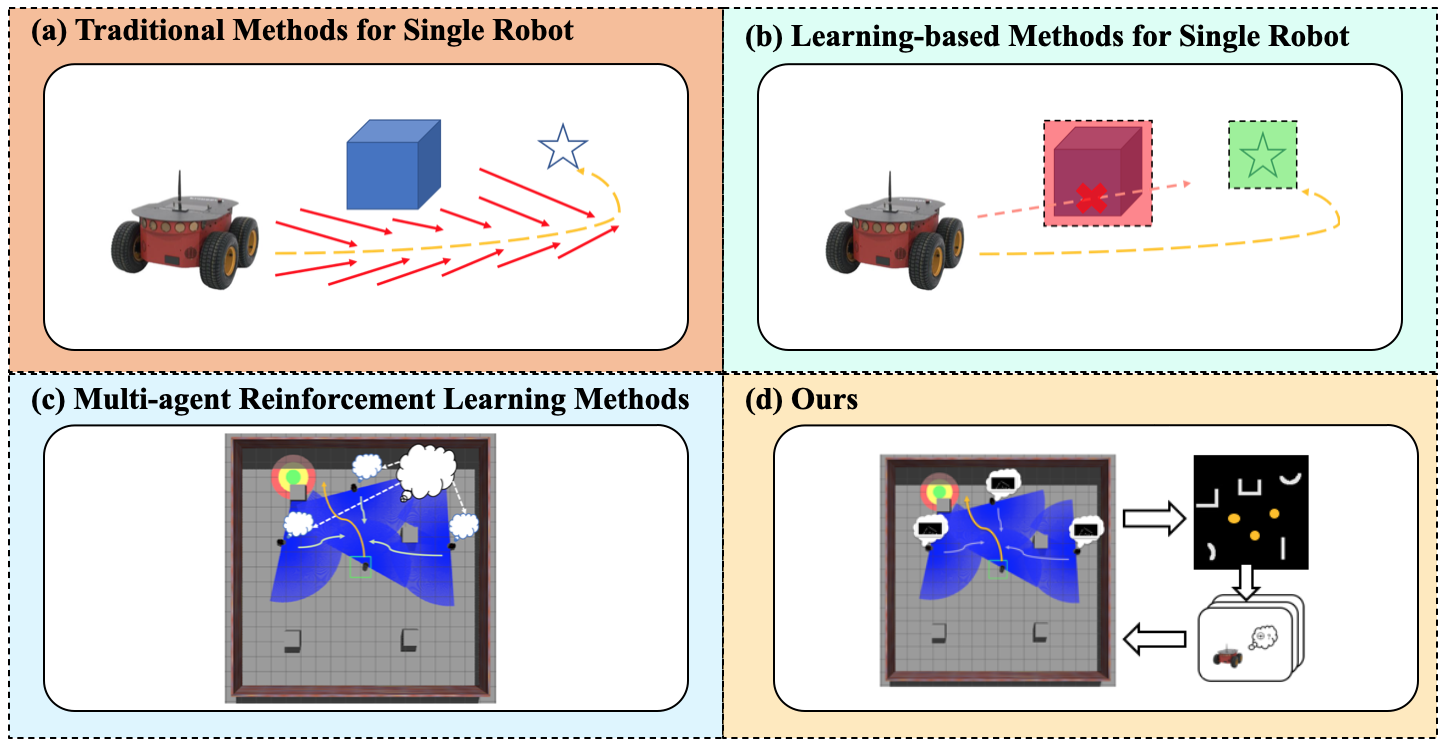}
\caption{\small{Illustration comparing traditional method for single robot(a), learning-based method for single method(b), multi-agent reinforcement learning method(c) and our method(d). Note that our method utilizes a potential field based module for formation control and a reinforcement learning policy using only 2D laser as a local planner.}}
\label{Comparison}
\end{figure}

Recently, researchers try to use multi-robot cooperation methods for improvement. Some of them directly transfer the single-robot methods to multi-robot settings by assigning different goals to each robot and executing path planning separately\cite{macalpine2015scram}\cite{adler2015efficient}. 
Other methods based on multi-agent reinforcement learning use global networks that have access to all agents' states and actions as guidance, leading the policy networks to learn cooperative behaviors\cite{khan2019learning}. 
However, these methods are only applicable when the number of robots is fixed and the separations between them are limited. And actual states of all the entities' states in the environment are usually needed in these methods, which can be quite unrealistic in the real world.

In this paper, we develop a multi-agent method for an arbitrary number of robots to follow the target in a socially-aware manner using only 2D laser scans, extending our previous work on social navigation task\cite{cui2020learning}. 
The framework of this method is composed of a potential field based module for formation control and a reinforcement learning policy network working as a local controller.
Specifically, the potential field module is responsible for figuring out the appropriate formation with an arbitrary number of robots according to the obstacle distribution around the target and assigning the positions to each robot. 
And the policy network navigates each robot to its own goal while avoiding collisions in a socially-aware manner.
The laser representation of sequential local obstacle maps disentangled from robots' ego-motion is used in these two modules to help the robots understand the environment, instead of requiring the actual states of all the objects in the environment or a pre-built global map.

Besides, our network is trained in a policy-sharing multi-agent simulation environment that better resembles actual crowds. The target and other robots move by sampling from the policy pre-trained with a socially aware navigation approach, rather than previous methods using simple predefined rules. 
Based on these designs, the learned policy can be directly transferred to an arbitrary number of robots and unseen environments in the person following task. 

Particularly, this paper presents the following contributions:
\begin{itemize}
	\item We propose a potential field based method for formation control and goal assignment for an arbitrary number of robots using a target-centered obstacle map.
	\item We propose a deep model of socially-aware following policy that navigates robots through crowded environments while following the target with only 2D laser scans.
	\item We train the policy in a decentralized policy-sharing multi-agent simulation environment that better resembles social interactions and actual target moving behaviors.
	
\end{itemize}

The paper is organized as follows: Section II is a brief summary of the related works. Section III describes the formulations and algorithm structure of our methods. Section IV shows the experimental results.

\section{Related Works}

\subsection{Single-robot Tracking with Traditional Methods}

Existing methods for single robot tracking tend to use a hierarchical pipeline that detects the target position first and navigates to it with traditional controllers like PID controllers or fuzzy controllers.

As for target detection and tracking, various sensors have been used in the person following task. Guevara et al. \cite{guevara2016vision} use RGB-D cameras to track specified targets with a TLD algorithm. Chen et al.\cite{chen2017person} use stereo cameras with an online Ada-Boosting algorithm to track the target human. And recent work of Cheng et al.\cite{8967645} have tried to use only web cameras to track both the target person's body and feet. While some other researchers try to use lightweight sensors like 2D laser scans in target following\cite{leigh2015person}
\cite{cosgun2013autonomous}, trackers like Kalman filter are used to locate the laser clusters as the target position.

After obtaining the target position, specific controllers are deployed to guide the robots to their own goal.
Morioka et al. \cite{morioka2004human} propose a control law based on the virtual spring model to realize relative tracking. Angus Leigh et al. \cite{leigh2015person} use PID controllers to minimize the position error while approaching the target position.
However, a problem with these methods is that the target may get lost for a variety of reasons like obstruction of obstacles. 
To address this problem, Lee et al. \cite{lee2018robust} use variational Bayesian techniques for trajectory prediction of the missed target and find the target again within a reasonable recovery time. 
Hoang et al. \cite{do2017reliable} propose a method in which the robot quickly navigates to the last position where the target is missed, and then use the Kalman filter to predict the possible target position if the target is not inside the field of view. 
Shin et al. \cite{shin2020optimization} develop an optimization-based method using the Following Field to maintain tracking and reduce recovery time at the same time. Following Field includes a repulsion field getting the robot out of the occluded area and an attraction field pushing the robot towards the target.

\subsection{Single-robot Tracking with Learning Methods}
Learning based approaches have also been applied to robot navigation tasks for social robots these years\cite{pang2020efficient}. These methods usually extract visual features from sensor data and directly map the features to the control command.

Imitation learning based ones learn the tracking policy from expert demonstrations. Those methods can work well in most cases but require a large number of training data for a specific task, which can be expansive in the real world. While reinforcement learning based ones learn from interacting with the environment, with the guidance from a reward function designed for the current task, instead of a human expert. However, those methods also need abundant interaction data for the policy training. Training in simulations can partially solve this problem, but the gap between the simulated sensors and real sensors like cameras still makes it hard to generalize to different environments.

\subsection{Multi-robot Planning}
Although there are ways to improve the performance of single-robot methods in many tasks, single-robot methods still have shortcomings such as limited field of view and lack of stability. To solve these problems, multi-robot methods are proposed these years.

Multi-robot navigation problem is usually decomposed into goal assignment problems and trajectory optimization problems. There are many graph based approaches that have been proposed to solve goal assignment problem\cite{yu2012distance} \cite{turpin2014goal}. However, those methods have some restrictions in initial conditions and dynamical approximations that limit their applicability. 

Closest to our work, Khan et al.\cite{khan2019learning} formulate the multi-robot tracking problem into a multi-agent reinforcement learning (MARL) problem, and solve goal assignment task and path planning task together. 
They use a global sparse reward function to update the centralized Q network, and then use the Q network to guide the training of each robots' policy network.
However, this framework is based on the knowledge of all the entities' states in the environment, which can be quite unrealistic in the real world. And this framework is applicable to an arbitrary number of robots and obstacles only in the training process meaning that it can not be applied to scenarios where robots keep joining in or getting lost. These drawbacks limit the flexibility and generality of this method.

\begin{figure}[t]
\centering
\includegraphics[scale=0.42]{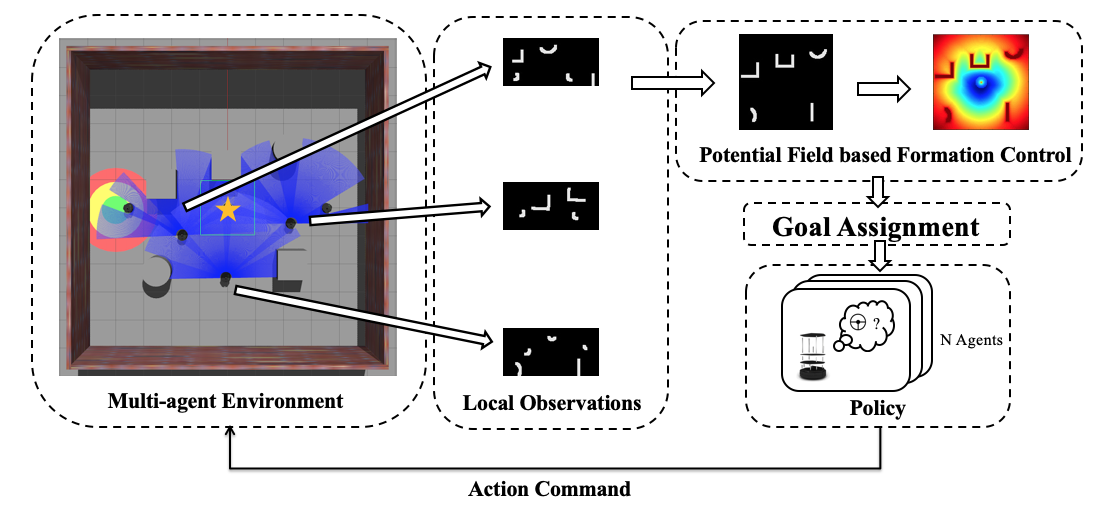}
\caption{\small{Overall framework. Each robot gets its own local observation in the multi-agent simulation environment. Then, the local observations get stitched together to form a target-centered obstacle map for potential field construction. With the help of the potential field, appropriate formation positions can be obtained and assigned to the robots. Finally, each robot navigates to its own goal by sampling action command from the learned policy.}}
\label{Framework}
\end{figure}

\section{Approach}

In this paper, we aim to tackle the multi-robot following problem where multiple robots navigate to approach and track the same target while avoiding collisions with each other and collisions with the surrounding obstacles in a socially compliant manner.
A reinforcement learning based planner is used to handle local interactions with both static and dynamic obstacles. And a potential field based formation control module is used to obtain the appropriate positions for an arbitrary number of following robots. 
Specifically, we use laser scans in sequential local obstacle map representation, which helps the robot understand the surrounding environments.
Besides, to enhance authenticity, we model the social interactions by using multiple following robots with shared policy and a target moving towards random goals with collision-free navigation algorithms. The following parts give details about the key ingredients of our framework.

\subsection{Deep Reinforcement Learning Settings}
The reinforcement learning based local planner is responsible for tracking a moving target while avoiding collisions with surrounding static and dynamic obstacles.
Both the moving patterns of the target and the obstacles should be considered in the decision making process.
Dynamic target following problem can be formulated as a Partially-Observable Markov Decision Process in reinforcement learning framework which can be described as a tuple $(\mathcal{S,A,P,R},\Omega,\mathcal{O})$, where $\mathcal{S}$ is the state space, $\mathcal{A}$ is the action space, $\mathcal{R}$ is the reward function, $\Omega$ is the observation space, $\mathcal{O}$ is the observation probability conditioned on the state and $\mathcal{P}$ is the state transition model of the environment dynamics. The detailed formulation is provided in the following part.

\textbf{Observation space}
Our observation space includes laser readings $o^t_l$, relative target position $o^t_t$ and robot's own velocity $o^t_v$ which can be directly acquired through sensors mounted on the robot and robot communication system.
And the laser readings $o^t_l$ are represented as obstacle maps.
Sequential obstacle maps are stacked by being transformed into the current frame based on the odometry, so that the static obstacles in scans are overlaid together, while the dynamic obstacles are highlighted, as shown in Fig. \ref{Framework}. In this way, the ego motion of the robot itself can be fully disentangled from the measurement, resulting in stacked maps making a clearer distinction between static and dynamic obstacles in the surrounding environment.

The relative target positions $o^t_t$ are recorded in chronological order within a specific time window to provide the target's moving pattern relative to each robot.

\textbf{Action space}
The action includes the linear and angular velocity of differential robots.
In the training and testing process, actions are sampled from policy $\pi$ in continuous space. While in the  initialization process, actions are sampled randomly from uniform distributions.
Considering the robots' kinematics, we set the range of linear velocity $v \in [0,0.7]$ and the angular velocity $w \in [-1.5, 1.5]$.
In order to reduce the difficulty, the target moves in a relatively lower linear velocity range of $v_t \in [0,0.4]$.

\textbf{Reward setup}
To guide the policy optimization, we have designed a reward function taking approaching target and collision avoidance into account:
\begin{equation}
R(s_t) = R_a(s_t) + R_c(s_t)
\end{equation}

Specifically, $R_a(s_t)$ is given to the robots for getting closer to its goal:
$$
R_a(s_t)=
\begin{cases}
r_{lost}&\text{if get lost}\\
w_1({\begin{Vmatrix} p^t-p_g^t\end{Vmatrix}-\begin{Vmatrix} p^{t-1}-p_g^{t-1}\end{Vmatrix}})&\text{else}\\
r_{arrive}&\text{if arrive}
\end{cases}
$$
where $p^t$ refers to the robot position at time step $t$, $p_g^t$ refers to the goal position at time step $t$. As we can see, this part of the reward is calculated based on the distance difference. Approaching the target will be rewarded with positive $R_a(s_t)$ and $r_{arrive}$ for arriving at the goal position, while moving away from the target will get penalized with negative $R_a(s_t)$. And if the robot gets too far away from the target position, we can assume that the robot has lost tracking and penalize it with $r_{lost}$.

$R_c(s_t)$ is set to penalize behaviors of getting close to the obstacles, making sure that the robots move in a safe manner. 

$$
R_c(s_t)=
\begin{cases}
r_{collision}&\text{if collision}\\
w_2(1-\cfrac{d}{r+r'})&\text{if ${d \leq r + r'}$} \\
0&\text{otherwise}
\end{cases}
$$
where $r$ here defines the safety radius of the robots, $r'$ defines a constant distance expanding the safety zones. $d$ refers to the minimum value of laser distance at the current time step and $w_2$ refers to a constant value. Temporal information encoded in the stacked local obstacle maps can help the policy network understand the moving patterns of the dynamic obstacles like pedestrians and react in advance, ensuring both self-safety and human comfort.

This reward setup will guide the robots to follow the moving target while paying attention to the surrounding objects.$w_1$ and $w_2$ are constants representing the importance of each factors.

\subsection{Multi-robot Simulation Environment}

To expose our local planner policy network to a simulation environment that resembles actual indoor following scenarios and test the performance of our overall framework, we adopt the idea of decentralized multi-agent setup proposed in \cite{fan2018fully} and build a cooperative environment in Gazebo. 
In this simulation environment, the robots are forced to learn cooperative behaviors when confronting each other.

In the local planner training process, all of the following robots share the same policy $\pi$ and make decisions by sampling from it without communication. All of them navigate to follow their own moving target without colliding with each other and the surrounding obstacles. The targets move in a fixed set of directions at a random velocity, avoiding over-fitting of the policy network.

In the overall performance testing process, all of the robots are asked to follow the same moving target and avoid collisions with each other. Each robot also makes decisions by sampling from the same policy $\pi$, while the goal positions for each robot in the formation is obtained from the potential field module and assigned to each one of them.
The target makes decisions according to a pre-trained socially aware navigation network\cite{cui2020learning} and moves to its own random generated goals.

Multiple scenarios are also designed in this simulation environment for both training and testing. All of the static obstacles and dynamic agents are randomized in sizes and shapes by setting collision ranges in Gazebo.
Both the following robots and the target are randomized in initialization with different positions and angles at the beginning of every episode. 
The number of following robots and the number of obstacles can also be changed, so that the robustness of the following framework can be tested.

\subsection{Policy network and Training Algorithm}

\begin{figure}[t]
\centering
\includegraphics[scale=0.3]{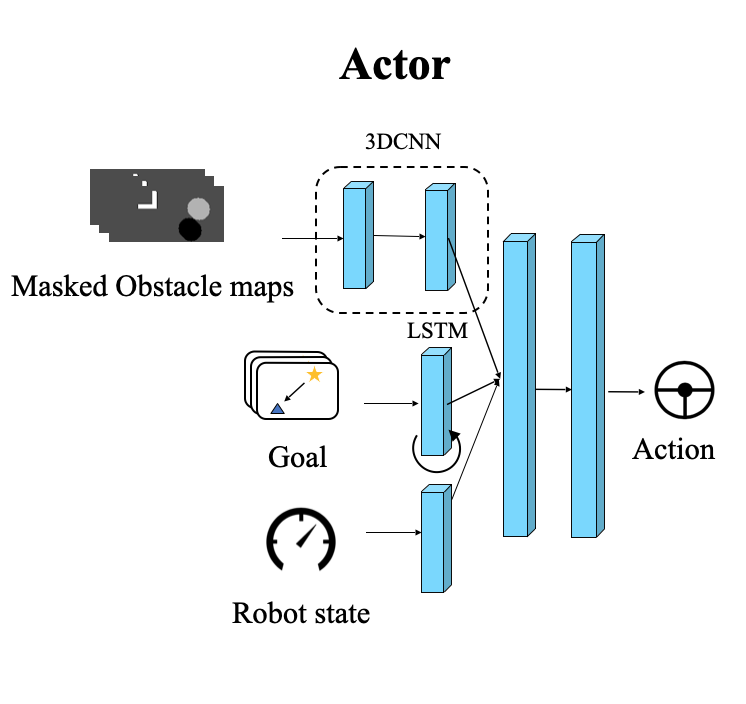}
\caption{\small{Architecture of actor network. It embeds state inputs and outputs action commands.}}
\label{actor}
\end{figure}

\begin{figure}[t]
\centering
\includegraphics[scale=0.3]{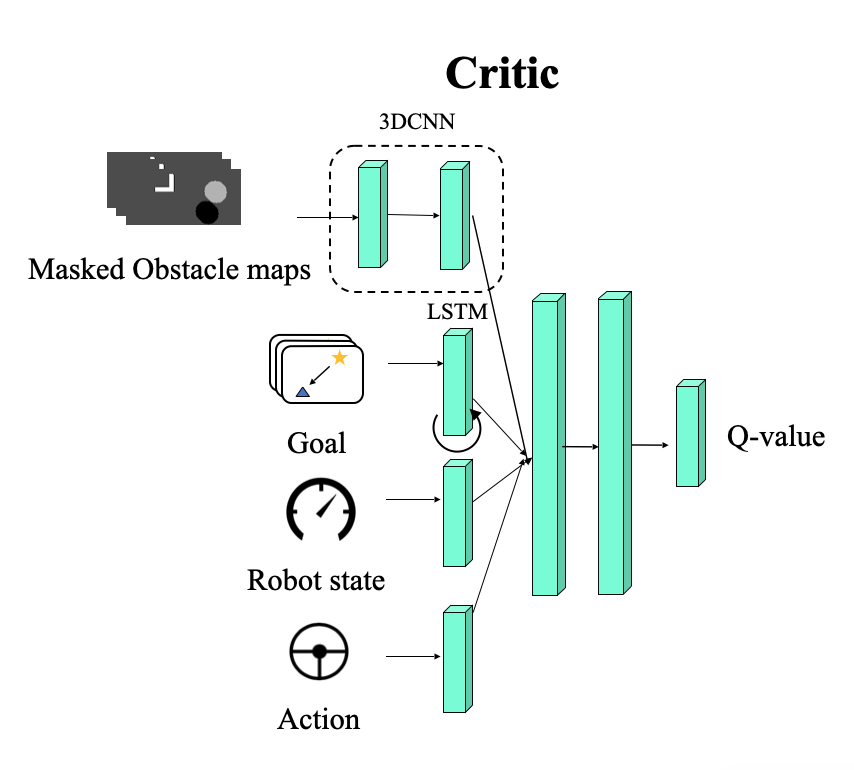}
\caption{\small{Architecture of critic network. It embeds states with corresponding actions and outputs Q-values.}}
\label{critic}
\end{figure}

As for the policy training network, we adopt the framework of TD3\cite{fujimoto2018addressing}. The actor network and critic network are presented in Fig. \ref{actor} and Fig. \ref{critic}. 
In both of the models, 3D-CNN modules are used to deal with the sequential obstacle map input and LSTM modules are used to encode the relative target positions in chronological order. 
All hidden states are concatenated together and further encoded by fully connected layers.
And all the state vectors are normalized to the range of $[0,1]$ before sending into the network.

The entire training process of the policy network is shown in Algorithm \ref{algorithm}. After random initial exploration, multiple robots make decisions independently but according to the same policy network. All of the robots are also used to collect interaction data in parallel, which speeds up the policy training. The policy network is trained until convergence.

\begin{algorithm}[tb]
\caption{Following policy training with deep reinforcement learning}
\label{algorithm}
\begin{algorithmic}[1]
\State Initialize policy $\pi_{\phi}$ and environment dataset $D_{env}$;
\For{E steps}
\State Take a random action for initial exploration;
\State Add real data to $D_{env}$
\EndFor
\For{N epochs}
\For{R steps}
\State Sample action $a_{i}$ from $\pi_{\phi}$ in real environment for robot $i$ according to its own state $s_{i}$;
\State Action $a_{i}$ executed independently by robot $i$;
\State Observe the next state ${s'}_{i}$ and reward $r_{i}$;
\State Add data $(s_{i}, a_{i}, {s'}_{i}, r_{i})$ to $D_{env}$
\For{P steps}
\State Update policy on environment data $\phi \leftarrow \phi - \lambda_{\pi} \hat{\nabla}_{\phi} J_{\pi}(\phi , D_{env})$;
\EndFor
\EndFor
\EndFor
\end{algorithmic}
\end{algorithm}

\subsection{Potential Field based Formation Control}

To obtain a formation that ensures both the safety of the robots and the following quality, we propose a potential field based formation control method.
The proper formation is obtained by acquiring a set of low-cost points from the potential field $F$ considering obstacle repulsion field $F_{ro}$, ally repulsion $F_{ra}$ and target attraction$F_{a}$:

\begin{equation}
F = F_{ro} + F_{ra} + F_a
\end{equation}

A target-centered obstacle map representing the surrounding space distributions is used to construct the obstacle repulsion field with only 2D laser readings, as shown in Fig. \ref{Framework}. The target-centered obstacle map is built by aggregating all the robots' laser readings and transforming them into the target's coordinate system. Both obstacles at the current time and historical obstacle trajectories are marked in this map, so that the temporal information of the environment can be recorded and the potential field can be calculated accordingly.
The repulsion field defined by $F_{ro}$ is in inverse proportion to the distance to the closest obstacles obtained from a Euclidean distance map algorithm.
And the target attraction field defined by $F_a$ is proportional to the second-order of the distance between the current position and the target position.
As acquiring a large number of robots' positions at the same time could be extremely computational inefficient.
We use an iterative method to calculate the formation positions in turn and update the potential field with the newly added robots' repulsion field defined by $F_{ra}$, which is similar to $F_{ro}$.
In this way, our method can generalize to an arbitrary number of robots with less time complexity. 

Then the positions in the formation are further assigned to the closest robot in the environment. Partial adjustment can be done locally to avoid path crossings that affect the efficiency of subsequent executions.

\section{Experiments}

\subsection{Formation Control with Potential Field}
\begin{figure}[t]
\centering
\includegraphics[scale=0.33]{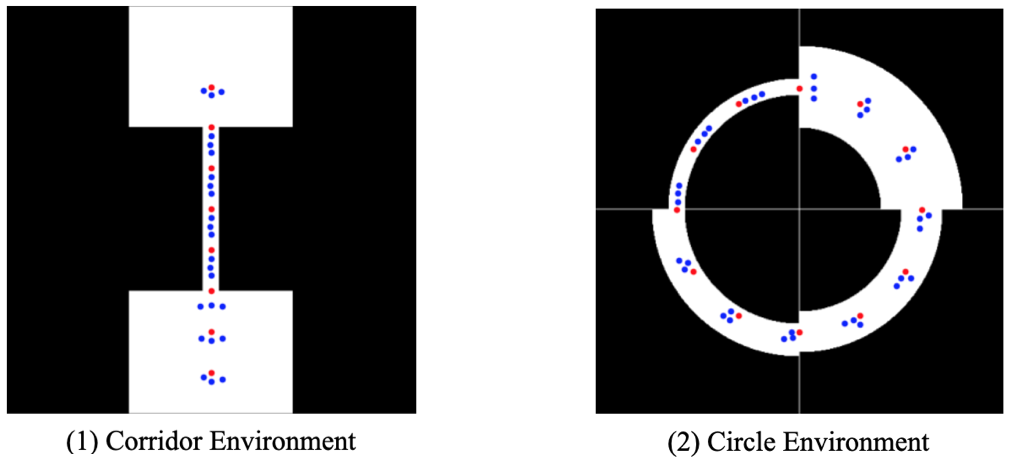}
\caption{\small{Formation planning in multiple environments. We test the performance of our potential field based formation planning method in multiple environments. Two typical scenes including corridor environment and circle environment are shown here. The red dots represent the target 's position, while the blue dots represent the following robots' positions. We can observe that the formation changes according to the local obstacle distributions, directing the robots to keep a distance from obstacles while following the target. }}
\label{multi_env}
\end{figure}

To prove the applicability of our potential field module, we test the formation planning performance in multiple scenarios, results are shown in Fig. \ref{multi_env}. 

We can see that the formation of the robots changes according to the shifting of the obstacle distributions around the target. The formation gets tightened to a line when the target gets through a narrow passage, while spreads out enough when the target stays in open areas. The robots also take the moving direction of the target into consideration and avoid getting in the way of the target.

Quantitatively, Tab. \ref{Scores} demonstrates that the potential field module substantially improves the safety performance of formation control compared with the fixed-position method in which target formation positions are set uniformly around the target ignoring the environment changes. With the help of this potential field module, the local planner can navigate the robot to an appropriate position, ensuring both safety and robustness of following.

\begin{table}[t]
    \centering
    \setlength{\tabcolsep}{6pt}
    \renewcommand{\arraystretch}{1.3}
    \begin{center}
        \label{PREDICTION}
        \begin{tabular}{l cc cc}
            \hline
            & \multicolumn{2}{c}{Corridor} & \multicolumn{2}{c}{Circle}  \\
            \cmidrule(r){2-3} \cmidrule(r){4-5} 
            & \makecell[c]{Average\\Distance} &
              \makecell[c]{Success\\Rate} & \makecell[c]{Average\\Distance} &
              \makecell[c]{Success\\Rate} \\
            \hline
            Fixed Positions         & 0.54m & 83.4$\%$ & 0.287m & 95$\%$ \\
            Potential Field         & \textbf{0.637m} & \textbf{100$\%$} & \textbf{0.386m} & \textbf{100$\%$} \\
            \hline
        \end{tabular}
        \caption{\small{Comparison of formation control strategy. We calculate the average distance between the robots and the nearest obstacles and the success rate of following the target in a fixed length of time, in our case, 30s. The potential field based method can achieve better performance.}}
        \label{Scores}
    \end{center}
    \vspace{-0.4cm}
\end{table}

\subsection{Local Planning with Reinforcement Learning}
To evaluate the performance of the local planner, we test the learned policy in multiple social scenarios, results are shown in Tab. \ref{social}.

Due to the limited field of view, the robots can obtain various laser readings even in simple randomized simulation environments. 
Therefore, the learned policy can directly generalize to unseen environments with an arbitrary number of dynamic obstacles represented with similar local readings.

\begin{table}[t]
    \centering
    \setlength{\tabcolsep}{6pt}
    \renewcommand{\arraystretch}{1.3}
    \begin{center}
        \label{PREDICTION}
        \begin{tabular}{l c c}
            \hline
                & Success Rate & Average Time \\
            \hline
            Passing         & 100$\%$ & 17.1s \\
            Crossing        & 95$\%$  & 15.8s \\
            Random          & 90$\%$  & 17.7s \\
            \hline
        \end{tabular}
        \caption{\small{Local planner evaluation. We test the learned policy's performance in multiple social scenarios. The robot navigates to catch a forward-moving target while avoiding collisions with robots coming from different directions in the process. Results show that the learned policy can achieve successful deployment in multiple unseen environments.}}
        \label{social}
    \end{center}
\end{table}

\subsection{Overall Framework Experiments}
To prove the merits of our multi-robot method, we compare the performance of it with a fixed-position multi-robot method and single-robot method on following robustness and safety assurance, results are shown in Tab. \ref{compare}.
Comparisons are conducted with the following metrics:
\begin{itemize}

\item \textbf{\emph{Following Scores:}} The ratio of steps keeping the target in the field of view and keeping a comfortable distance from the target. Let m be the steps keeping following and N be total steps, $\emph{Following Score} =  m /N * 100$.

\item \textbf{\emph{Average Distance:}} Average distance to the closest obstacle in the following process.

\end{itemize}

Fixed-position multi-robot method assigns the formation points to robots without paying attention to the surrounding environment of the target, usually leading to occlusions when the target formation intersects obstacle areas. While the single-robot method tends to lose tracking in crowded environments due to the limitation of vision.
In comparison, our method shows substantial improvement of the following robustness and a proper degree of safety awareness in the evaluation. 

\begin{table}[t]
    \centering
    \setlength{\tabcolsep}{6pt}
    \renewcommand{\arraystretch}{1.3}
    \begin{center}
        \label{PREDICTION}
        \begin{tabular}{l c c}
            \hline
                & Following Scores & Average Distance \\
            \hline
            Single-robot         & 57.9$\%$ & \textbf{1.25m}\\
            Fixed-position Multi-robot        & 75.1$\%$  & 0.826m \\
            Ours          & \textbf{85.3$\%$}  & 1.05m \\
            \hline
        \end{tabular}
        \caption{\small{Overall framework evaluation. We compare the performance of our method with the fixed-position multi-robot method and the single-robot method. Results show that our method can achieve performance with better robustness and safety awareness.}}
        \label{compare}
    \end{center}
\end{table}

\subsection{Real World Experiments}
\begin{figure}[t]
\centering
\includegraphics[scale=0.36]{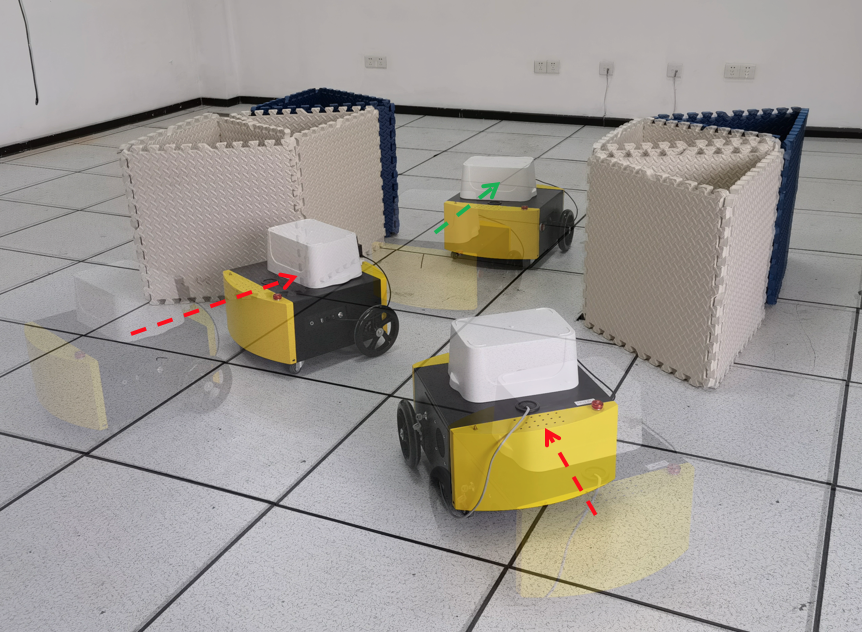}
\caption{\small{Real robot evaluation. Figures above show the trajectories of real robots following the target going through a narrow corridor. Robots tighten up their formation to go through the corridor. The green and red arrows mark the trajectories of the target robot and the following robots respectively.}}
\label{real}
\end{figure}

We also evaluate our method on real differential robots, example result of the team of two robots following the target through a narrow corridor is shown in Fig. \ref{real}. The robots can change their formations to adapt to the target-centered environments, so that both the self-safety of the robots and the following stability can be guaranteed.

\section{Conclusions}

This paper presents a multi-agent method for an arbitrary number of interchangeable robots to follow the target in a socially-aware manner using only 2D laser scans.
By utilizing potential field and reinforcement learning, we can navigate a team of robots through unseen environments while following the target in multiple dynamic environments, ensuring both safety and robust following performance.Future work will focus on adapting our method to complex environments that need formation splitting and merging.

\section{Acknowledgement}
This work was supported by the Science and Technology Project of Zhejiang Province (2019C01043).



\bibliographystyle{ieeetr}

\end{document}